\documentclass{article}

\usepackage{PRIMEarxiv}

\usepackage[utf8]{inputenc} 
\usepackage[T1]{fontenc}    
\usepackage{hyperref}       
\usepackage{url}            
\usepackage{booktabs}       
\usepackage{amsmath}
\usepackage{amsfonts}       
\usepackage{nicefrac}       
\usepackage{microtype}      
\usepackage{lipsum}
\usepackage{fancyhdr}       
\usepackage{graphicx}       
\graphicspath{{media/}}     
\usepackage[ruled]{algorithm2e}
\usepackage{multirow} 
\usepackage{threeparttable}
\usepackage{bm} 
\usepackage{subcaption}
\usepackage{graphicx}
\usepackage{thmtools}
\usepackage{thm-restate}
\usepackage{cleveref}
\usepackage{amsthm}

\newtheorem{theorem}{Theorem}[section]
\newtheorem{lemma}[theorem]{Lemma}

\newtheorem{corollary}[theorem]{Corollary}
\newtheorem{definition}{Definition}[section]

\title{Federated Graph-based Networks with Shared Embedding
}

\author{
  Tianyi Yu\\
  School of Electrical and Information Engineering\\
  Beijing University of Civil Engineering and Architecture \\
  Beijing\\
  \texttt{201906020144@stu.bucea.edu.cn} \\
  \And
  Pei Lai\\
  School of Information Science and Technology \\
  Southwest Jiaotong University\\
  Chengdu\\
  \texttt{peilai@my.swjtu.edu.cn} \\
   \And
  Fei Teng\\
  School of Information Science and Technology \\
  Southwest Jiaotong University\\
  Chengdu\\
  \texttt{fteng@swjtu.edu.cn} \\
}

\begin{document}
\maketitle

\begin{abstract}
Nowadays, user privacy is becoming an issue that cannot be bypassed for system developers, especially for that of web applications where data can be easily transferred through internet. Thankfully, federated learning proposes an innovative method to train models with distributed devices while data are kept in local storage. However, unlike general neural networks, although graph-based networks have achieved great success in classification tasks and advanced recommendation system, its high performance relies on the rich context provided by a graph structure, which is vulnerable when data attributes are incomplete. Therefore, the latter becomes a realistic problem when implementing federated learning for graph-based networks. Knowing that data embedding is a representation in a different space, we propose our Federated Graph-based Networks with Shared Embedding (Feras), which uses shared embedding data to train the network and avoids the direct sharing of original data. A solid theoretical proof of the convergence of Feras is given in this work. Experiments on different datasets (PPI, Flickr, Reddit) are conducted to show the efficiency of Feras for centralized learning. Finally, Feras enables the training of current graph-based models in the federated learning framework for privacy concern.
\end{abstract}


\section{Introduction}
Recently, the user privacy issue on  web applications arouses lots of attention. With the development of network technology and improvement of IoT, a large volume of personal data is collected and analysed in daily life and some of them may be uploaded to online network illegally. In 2018, the European Union (EU) and the European Economic Area (EEA) enacted the General Data Protection Regulation (GDPR) \cite{EUdataregulations2018}, whose main purpose is to protect data privacy and to limit personal data transfer between processors.

The launch of GDPR is an epitome of public concern on privacy in a digital age. In fact, the discussion on privacy protection techniques can be traced back to decades ago. \cite{privacyprotec} explains that the main problem is not due to the lack of available security mechanisms but the privacy preserving on computational aspect. Several technologies are brought forward today: trusted execution environment \cite{7345265}, secure multiparty computation \cite{4568388,4568207,9183133}, federated learning \cite{Li_2020, yang2019federated}. Among these three mechanisms, federated learning, which employs multiple devices to build a model collaboratively while keeping all the data in local storage, stands out because of its preeminence of low computational costs. What's more, federated learning also borrows tools such as differential privacy \cite{wei2019federated}, secure multiparty computation \cite{4568388}, homomorphic encryption \cite{acar2017survey} to offer enhanced security.  

Graph Neural Networks (GNN) and their variants are showing dominant performance on many ML tasks \cite{zhou2021graph}. For example, Graph Convolutional Network (GCN) has been widely applied in many different domains such as semi-supervised classification \cite{kipf2017semisupervised}, multi-view networks \cite{khan2019multigcn} and link prediction \cite{zhang2018link}, etc.  However, the efficiency of graph-based networks relies on the rich information among the elements contained in a graph. Once faced with a perturbed topological structure or incomplete node attribute, the network either can no longer provide satisfied learning performance \cite{Z_gner_2018} or needs a complex node attribute completion system \cite{Chen_2020}. The nature of graph-based networks results in the hard implementation of federated learning, especially when the transmission of original user data, such as features and labels of nodes, is prohibitive between processors.

Based on the observation that the embedding representation differs from initial data space, we propose our \underline{Fe}derated G\underline{r}aph-b\underline{a}sed Networks with \underline{S}hared Embedding (\textbf{Feras}) model in this paper. Instead of exchanging raw data information at the beginning, we facilitate embedding communication in the middle of the network. On the one hand, computations are performed on edge hosts synchronously to speed up training without leaking raw data, on the other hand, node information is propagated through the entire graph via embedding, which is critical to retain the network performance.

The rest of the paper proceeds as follows: we begin by discussing the latest related research of distributed learning and training methods, then go introducing four main rules in application scenarios and proposing algorithm structure. The next section is mainly concerned with the theoretical analysis and experiment results are reported in the following. Some concluding remarks are put in the end.

\section{Background}
\subsection{Fedeated learnig}
Federated learning was first proposed by Google \cite{mcmahan2017communicationefficient} in 2016 for Android mobiles phones. Its main idea is to train centralized model with decentralized data. Indeed, devices use local data to train models and upload parameters to server for aggregation. One of the most important features of federated learning is security and privacy \cite{yang2019federated}. Even though sensitive data are not directly shared between devices, it is still urgent to protect the communication between server and device. Represented by homomorphic encryption \cite{Rivest1978,Hall2011SecureML,10.1007/978-3-319-93387-0_13}  and secure multiparty computation \cite{Mohassel2017SecureMLAS,kilbertus2018blind}, encryption methods provide a very safe and reliable solution while their computation cost is relatively high. Perturbation methods, such as differential privacy \cite{geyer2018differentially,mcmahan2018learning}, use a designed noise mechanism to protect sensitive information, while no extra computation cost is demanded, there may exist risks on prediction accuracy. However, information can still be leaked indirectly with intermediate results from updating process, \cite{wainakh2021user, melis2018exploiting} exploit some attacks and provide defense mechanisms.

Most recently, a large scale of work has been done to implement federated learning in graph-based networks. \cite{wang2020graphfl} addresses non-iid issue in graph data and confronts tasks with new label domains. \cite{he2021spreadgnn} develops a multi-task federated learning method without a server. \cite{wu2021fedgnn} suggests FedGNN, a recommendation federated system based on GNN. \cite{zhang2021subgraph} proposes FedSAGE, which is a sub-graph level federated learning technique utilizing graph mining model GraphSAGE \cite{hamilton2018inductive}. \cite{meng2021crossnode} introduces a federated spatio-temporal model to improve the forecasting capacity of GNN. Nevertheless, to the best of our knowledge, there is still no study concentrating on node embedding aggregation on sampling-based graph neural network framework. 

\subsection{Related work}
GraphSAINT \cite{zeng2020graphsaint} adopts an innovative way to build mini-batches. Instead of sampling across the layers, GraphSAINT samples sub-graphs at first and then constructs a full GCN on it, which prevents neighbor explosion problem. It also establishes an algorithm to reduce sampling bias and variance. From comparison experiments conducted on datasets in \cite{zeng2020graphsaint}, GraphSAINT shows absolute advantages on accuracy and convergence rate compared with other popular samplers (GraphSAGE \cite{hamilton2018inductive}, FastGCN \cite{chen2018fastgcn}, ClusterGCN \cite{2019clusterGCN}, etc).

A stochastic shared embedding (SSE) for graph-based networks is investigated in \cite{wu2020stochastic} to mitigate over-parameterization problems. Embeddings are trainable vectors that are exchanged according to transition probability for each backpropagate step. Our proposed \textbf{Feras} is very different from SSE since \cite{wu2020stochastic} does not consider privacy issues and runs all computations on the same server.

To prove the convergence of parallelized stochastic gradient descent, \cite{NIPS2010_abea47ba} offers a novel idea by borrowing the good convergent property of contraction mappings, which inspired the mathematical demonstration of our paper even though proof in \cite{NIPS2010_abea47ba} is drawn for traditional neural network and no embedding sharing mechanism is acknowledged.

\section{Proposed Approach}
\subsection{An alternative of federated learning}
In this paper, we attempt to fill in the gap between privacy protection and traditional graph-based network. Let's begin with a simple user privacy scenario. A social network is built by integrating the client network of different companies, the global topology structure is thus shared between each company. However, because of the sensitivity of user information, personal data may not be shared directly between each company. For a client who buys the service of several companies at the same time, its information is accessible to all seller companies. In such cases, a user can be represented by a node and its features are only visible to its host company. 

\begin{figure}[h]
  \centering
  \includegraphics[width=\linewidth]{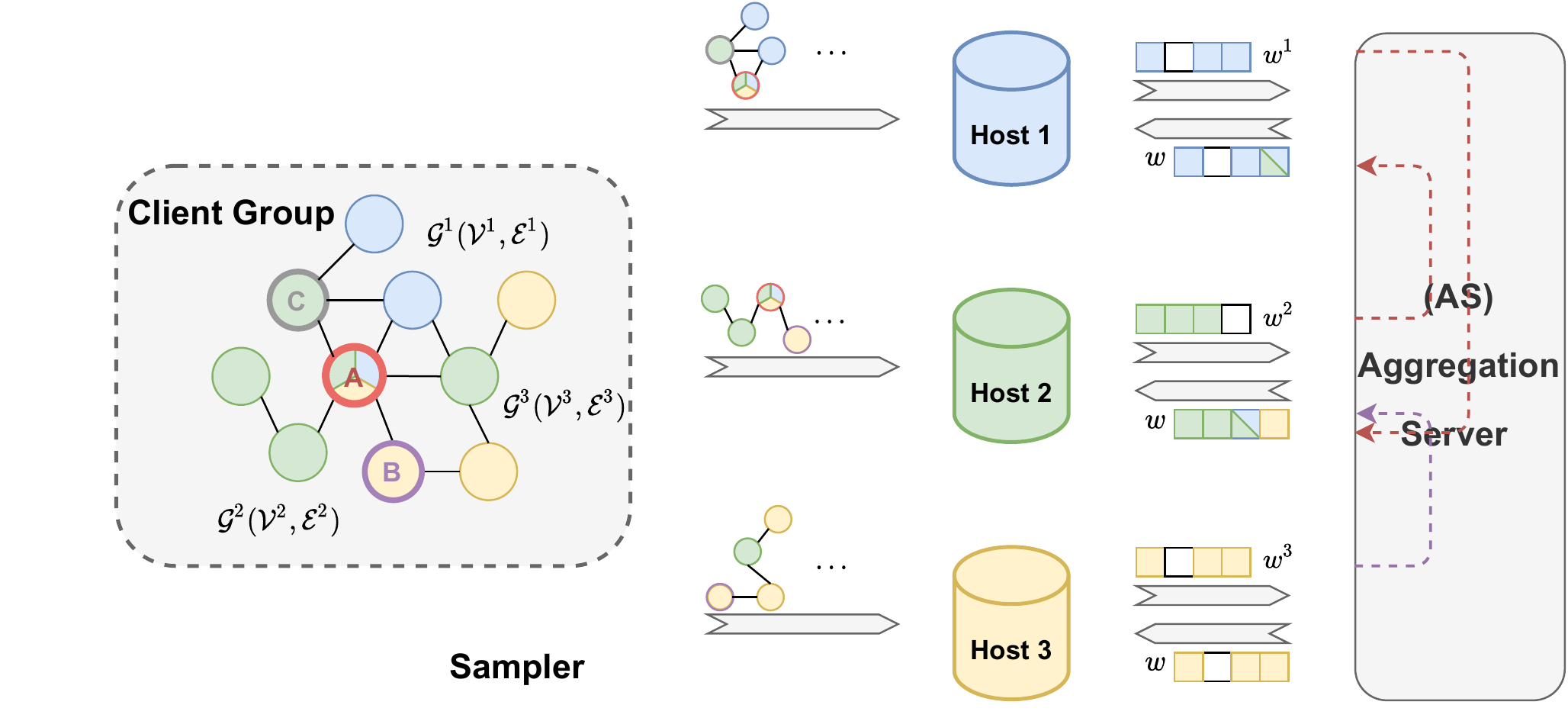}
  \caption{Relation between Client, Sampler, Host and AS. Hosts can only see nodes with the same color as their own. Training parameter is denoted by $w$. For embedding of an unseen node, hosts push a blank vector to AS and pull the latest overall embedding from it. The propagation of embeddings is manifested with a dotted line in AS.}
  \label{approach}
\end{figure}

While traditional GCN training on separate host machines is unable to prevent the network from accuracy decrease due to the lack of information, our \textbf{Feras}  model still allows companies to corporate graph training via an Aggregation Server (AS) without sharing the raw data. Turning now to a general privacy problem, four main roles can be retrieved as below and their relation is depicted in Figure \ref{approach}:

\textbf{Client: }A client is a basic element of the network, it can be interpreted as a held node. Besides information (attributes) attached to each client, which is the input of the network, it also records which host it belongs to. Two types of nodes are classified: private node (visible to only one host), and public node (visible to two or more hosts).

\textbf{Host: }Each host $n$ only has access to a certain number of nodes $H(n)$ and graph topology. For a distributed sub-graph $\mathcal G^{n}(\mathcal V^{n},\mathcal E^{n})$, host $n$ is capable to launch the training program with its available information. What's more, all vectors associated with unseen nodes in $\mathcal V^n$ are initialized as zero. Hosts need to communicate with AS to gain more knowledge on the network.

\textbf{Sampler: }The sampler is a work distributor. It is supposed to divide the entire network into sub-graphs and distribute them to each host. The sampler should eliminate sampling bias and improve training efficiency as much as possible.

\textbf{Aggregation Server (AS): }The AS works as a bond in our model. It collects the latest information from each host and gives them back after treatment. The latest information can be anything except raw data, such as embeddings, gradient, or weight matrices. To improve network latency, weight matrices can be shared after every $q$ iterations $(q\in \mathbb{N}^*)$. 

\subsection{Learning algorithm}
In this paper, we focus on node classification using convolutional layers. The latest GraphSAINT \cite{zeng2020graphsaint} is adopted as the sampler and the AS is more concerned with sharing embeddings and weight matrices. 

Information of node $v$ is recorded as input $x_v$ and label $y_v$. For a GCN of $m$ layers, the embedding is set to be shared right after $p$-th layer$(1\leq p\leq m)$. We note $\mathbb E(\mathcal V^n)$ the embedding set of node group $\mathcal V^n$ given to host $n$ and recall that $H(n)$ denotes the set of visible nodes to host $n$. The learning algorithm is presented in Algorithm \ref{algo1}.

Generally speaking, for host $n$ at iteration $t$, the sub-graph accepted from sampler $\mathcal G^{n,t}(\mathcal V^{n,t},\mathcal E^{n,t})$ may contain unseen nodes, the first step is to set the attribute of these nodes to zero, then feed the first $p$-layer network to generate an embedding for each node. So far all hosts have worked on their own in parallel and pushed node embedding to the AS. The sharing mechanism of AS is set to be an average compute between proprietaries, which means that the embedding of a given node depends exclusively on its host(s), examples of sharing mechanism are given later. 

After receiving the latest embeddings from AS, hosts use them to feed the rest $m-p$ layers and update weight matrices with back propagation. Resembling node embeddings, weight matrices are also pushed to AS and are pulled later by hosts after the average computing in AS.

\begin{algorithm}
\LinesNumbered
    \SetKwInOut{Input}{Output}
    \KwIn{Training Graph $\mathcal G(\mathcal V,\mathcal E)$, $m$ layer network, shared-embedding index $p$, input $x_v$, label $y_v$, $N$ host, iteration $T$}
    \KwOut{Parameter $w_i$}
    
    Initialize $w$
    
    \For{all $t\in\{1,\cdots,T\}$}{
    $\mathcal G^{n,t}(\mathcal{V}^{n,t},\mathcal{E}^{n,t})\leftarrow$ Sampled sub-graphs from $\mathcal{G}(\mathcal{V},\mathcal{E})$
    \For{all $n\in\{1,\cdots,N\}$ \textbf{parallel}}{
    $w^{n,t}\leftarrow w$
    
    $x_v\leftarrow 0$ for $v\in\{\mathcal V^{n,t}\setminus H(n)\}$
    
   $\mathbb E(\mathcal{V}^{n,t})\leftarrow$ Forward propagation after $p$ layers
    
    Push $\mathbb E(\mathcal{V}^{n,t})$ to Aggregation Server
    }
    Pull $\mathbb E(\mathcal V^{t})$ from Aggregation Server 
    
    \For{all $n\in\{1,\cdots,N\}$ \textbf{parallel}}{
    $\mathbb E(\mathcal V^{n,t})$ pulled from $\mathbb E(\mathcal V^{t})$
    
    $\hat{y}_v\leftarrow$ Forward propagation after $m$ layers, $v\in\mathcal V^{n,t}$
    
    $w^{n,t}\leftarrow$ Back propagation
    
    Push $w^{n,t}$ to Aggregation Server
    }
    
    $w\leftarrow AS(w^{1,t},\cdots,w^{N,t})$

    }
    
    \caption{Federated Graph-based Networks with Shared Embedding (Feras) algorithm}
    \label{algo1}
\end{algorithm}

Now we further clarify how AS determines the node representation in line 9 according to embeddings received from all hosts. During iteration illustrated in Figure \ref{approach}, node A is visible to all hosts but distributed to only two hosts, thus its representation is an average of embedding obtained by host 1 and host 2. For node B, even though it's a private node of host 3, its embedding is still accessible through AS for other hosts. Differently, for private node C, its embedding is not available in AS because it's not sampled by its proprietary host 2.

Similarly, in line 16 at the end of each iteration, the parameter given back to each host via AS is the average of all parameters calculated by different hosts. 

\section{Theoretical Guarantee}
\label{theory}
Despite the ideology behind Algorithm \ref{algo1} being intuitively legible, a rigorous convergence analysis is given in this part with more detailed proof in the appendix. 
\begin{figure}[h]
  \centering
  \includegraphics[width=\linewidth]{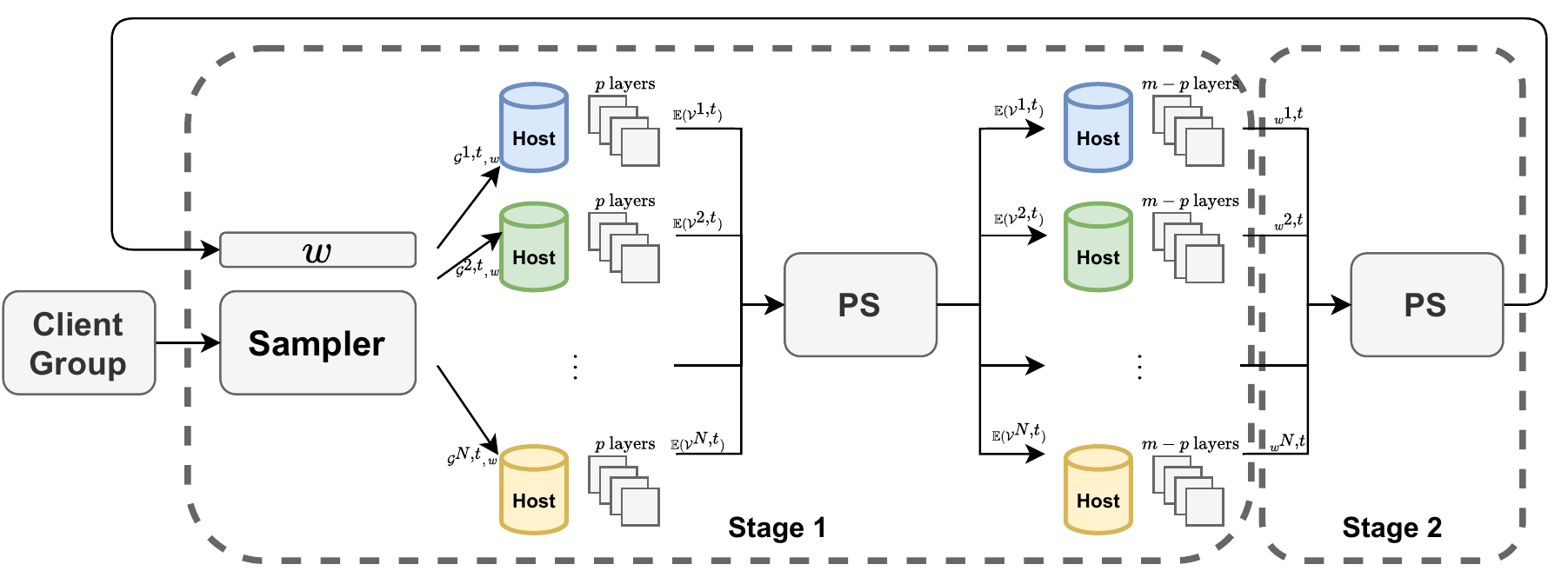}
  \caption{An illustration of training process}
  \label{stages}
\end{figure}
Note that the core idea of our model can be split into two main stages as showed in Figure \ref{stages}: a feed-forward network with shared-embedding layers and the back forward update followed by an  average-weight-matrix compute. We can thus outline our proof strategy as follows :
\begin{itemize}
    \item Formulate a representative application of our model with well-defined parameters.
    \item Explore the convergence property. Given a fixed learning rate $\eta$, the update rule for each host guarantees the convergence of weight matrices. In regard to the second stage, averaging the weight matrices of each host can be considered as a merge of parallelized stochastic gradient descent algorithm and a contraction mapping rule. This results from the fact that each host draws their data from the same data distribution and only uses a random part of data for gradient descent,  while benefiting from the whole dataset embedding in their feed forward network.
    \item Analyze the convergence rate and accuracy. The shared-embedding layer is actually a share of information, which contributes to a looser parameter constraint for a given convergence rate. 
\end{itemize}

\subsection{Preliminaries}
Without loss of generality, we implement our model to a simple scenario with a two GCN layer ($L_1$ and $L_2$) and multiple $N$ hosts. For $i\in\{1,2\}$, host $n$, and iteration $t$, the linear operator is identified as $w_i^{n,t}$ and the non-expansive averaged activation function as $\sigma_i$. The share-embedding method is modeled by a layer $\mathcal S$ between layer $L_1$ and $L_2$.

For each iteration, let $A^{n,t}$ be the adjacency matrix of $\mathcal{G}^{n,t}$ and $k_{n,t}$ its dimension.  At iteration $t$, note $\mathcal V^{n,t}=[ v^{n,t}_1,\cdots,v^{n,t}_{k_{n,t}}]$ the list of nodes distributed to host $n$ and $\mathcal V^{t}=[\mathcal V^{1,t},\mathcal V^{2,t},\cdots,\mathcal V^{N,t}]$ the concatenate of all $\mathcal V^{n,t}$. It is clear that :
\begin{equation}
    K_t=len(\mathcal V^t)=\sum_{n=1}^N k_{n,t}
\end{equation}
Define $\widetilde A^{n,t} = A^{n,t}+I_{k_n}$, $\widetilde D^{n,t}_{ii}=\sum_j \widetilde A^{n,t}_{i,j}$, $\overline A^{n,t}=(\widetilde D^{n,t})^{-\frac{1}{2}}\widetilde A^{n,t}(\widetilde D^{n,t})^{-\frac{1}{2}}$. For all $ n\in \{1,\cdots,N\},t\in\{1,\cdots,T\}$, recalling that $x^{n,t}$ is the input of graph, the feed-forward network can be summarized as :
\begin{align} 
Layer \quad L_1: \quad\hat x^{n,t}=&\sigma_1(\overline A^{n,t}x^{n,t}w_1^{n,t})\label{hat x}\\ 
Layer \quad \mathcal{S}:\quad\widetilde x^{n,t}=&S(\hat x^{n,t})\label{widetilde x}\\
Layer \quad L_2:\quad\hat y^{n,t}=&\sigma_2(\overline A^{n,t}\widetilde x^{n,t}w_2^{n,t})
\label{hat y}
\end{align}
where 
\begin{align} 
&x^{n,t}\in \mathbb R^{k_{n,t}\times m_1},\hat x^{n,t},\widetilde x^{n,t}\in\mathbb R^{k_{n,t}\times m_2},\hat y^{n,t}\in\mathbb R^{k_{n,t}\times m_3}, \\ 
&w_1^{n,t}\in \mathbb R^{m_1\times m_2},w_2^{n,t}\in \mathbb R^{m_2\times m_3}.\label{def space}
\end{align}
Recalling that $H(n)$ is node set visible to host $n$, define $\widetilde{\mathcal{V}}^{n,t}=[h^n(v_1^{n,t}),\cdots,h^n(v_{k_{n,t}}^{n,t})]$ where $h^n(v)=v1_{H(n)}(v)$ and $1$ is the indicator function. Similarly, we define $\widetilde{\mathcal V}^{t}=[\widetilde{\mathcal V}^{1,t},\widetilde{\mathcal V}^{2,t},\cdots,\widetilde{\mathcal V}^{N,t}]$. We can then further give an explicit expression of $S(\cdot)$:
\begin{equation}
    \widetilde X^{n,t}=\Theta^{n,t}\hat{x}^t
    \label{connect equa}
\end{equation}
where $\Theta^{n,t} =(\theta^{n,t}_{i,j})_{1\leq i\leq k_{n,t},1\leq j\leq K_t}$ representing how the information is propagated through the AS and $\hat{x}^t$ is the stack of all the intermediate output $\hat{x}^{n,t}$:
\begin{align}
    &\theta_{i,j}^{n,t}=\begin{cases}\frac{1}{ \widetilde{\mathcal V}^t.count(v_i^{n,t})} &\quad \text{ if } v^{n,t}_i=\widetilde{\mathcal V}^t(j)\\ 0 &\quad \text{ if }v^{n,t}_i\neq \widetilde{\mathcal V}^t(j)\end{cases}
\end{align}
\begin{align}
    \hat{x}^t &=(\hat{x}^{1,t},\cdots,\hat{x}^{N,t})^T\\
    &=(\hat{x}^{1,t}(v^{1,t}_{1}),\cdots,\hat{x}^{1,t}(v^{1,t}_{k_{1,t}}),\cdots,\hat{x}^{N,t}(v^{N,t}_{1}),\cdots,\hat{x}^{N,t}(v^{N,t}_{k_{N,t}}))^T\notag
\end{align}

In this paper, we limit ourselves to convex loss function $c^{n,t}:l_2\to [0,\infty)$ and the regularized risk minimization is stated as :
\begin{equation}
    c^{n,t}(w^{n,t})=\frac{\lambda}{2}\|w^{n,t}\|^2+L(x^{n,t},y^{n,t},\hat y^{n,t},\widetilde x^{n,t})
    \label{cost function}
\end{equation}
where $w^{n,t}=\{w_1^{n,t},w_2^{n,t}\}$, $L$ is the loss between $y^{n,t}$ and $\hat y^{n,t}$ which is convex. Applying stochastic gradient descent mapping :
\begin{equation}
    \label{update rule}
    \forall i\in\{1,2\},\widetilde w^{n,t+1}_i\leftarrow \phi^{n,t}:=w^{n,t}_i-\eta\nabla c^{n,t}(w_i^{n,t})
\end{equation}
The training process can be considered as a multi-machine parallel gradient descent. At the end of each iteration, each host pushes their weight matrices to AS and pulls the latest weight matrix as follows :
\begin{equation}
    w^{n,t+1}_i=\sum_{n=1}^N \frac{\widetilde w^{n,t}_i}{N}
\end{equation}
Now our goal is established to prove that the averaged model trained by each edge host can minimize the global cost function:
\begin{equation}
    \forall i\in\{1,2\},C(w)=\sum_n c^{n,T}(w^{n,T})/N
\end{equation}

\subsection{Contraction and convergence}
\label{contraction and convergence}
Before delving into details, let's start by some basic definitions and the well established Banach fixed-point theorem \cite{Latif2014} :
\begin{definition}[Lipschitz continuity]
\label{Lipschitz continuity}
A function $f:\mathcal X\to \mathbb R$ is Lipschitz continuous with constant $L$ with respect to a distance $d$ if $|f(x)-f(y)|\leq Ld(x,y), \forall x,y \in \mathcal X$.
\end{definition}
\begin{definition}[Contraction]
\label{Contraction}
For a metric space $(M,d)$, $f:M\to M$ is a contraction mapping if $\|f\|_{Lip}\leq 1$ where $\|f\|_{Lip}$ is the smallest constant for which the Lipschitz continuity holds.
\end{definition}
\begin{theorem}[Banachs Fixed Point Theorem]
\label{fixed point theorem}
If $(M,d)$ is a non-empty complete metric space,then any contraction mapping $f$ on $(M,d)$ has a unique fixed point $x^*=f(x^*)$.
\end{theorem}

Expression (\ref{def space}) shows that our parameter definition space is a metric space. Inspired by Theorem \ref{fixed point theorem}, our analysis is driven by the idea that the update rule (\ref{update rule}) is a contraction.

\begin{restatable}[]{lemma}{restatea}
\label{lemma1}
Given cost function $c(w)=\frac{\lambda}{2}\|{w}\|^2+L(x,y,\hat y,\widetilde x)$, $u(\cdot)$ the derivative function of $L$ over $\hat{y}$, $u(\cdot)$ is Lipschitz continous with constant $c^*$. If learning rate $\eta$ and element-wise bound is small enough, the update rule  $\phi(w^n):=w^{n}-\eta\nabla c(w^n)$ is also Lipschitz continuous with constant $1-\frac{\eta\lambda}{2}$ and is thus a contraction.
\end{restatable}
The detailed proof is given in Appendix \ref{appendix A}. Lemma \ref{lemma1} clarifies the contraction property of feed forward network in the first stage of Figure \ref{stages}, now let's turn to the weight share process. 

\begin{lemma}
\label{lemma2}
Given a Radon Space $(M,d)$, if $p_1,p_2,\cdots,p_n$ are contraction mappings with constants $c_1,c_2,\cdots,c_n$ with respect to the Wasserstein distance $W_z$, and $\sum_ia_i=1$ where $a_i\geq 0$, then $p=\sum_{n=1}^N a_ip_n$ is a contraction mapping with a constant of no more than $[\sum_i a_i(c_i)^z]^\frac{1}{z}$. Specifically, if for all $i$, $c_i\leq c$, then the contraction constant of $p$ is no more than $c$.
\end{lemma}

This is proven in paper \cite{NIPS2010_abea47ba}. We apply it to the average operation: define $p_n=\phi^n(w^n)$, $p^*=\sum_{n=1}^N a_i\phi^n(w^n)$, since $p_n$ is a contraction with constant $1-\frac{\eta\lambda}{2}$, $a_i=1/N$,  $p^*$ is also a contraction with constant $1-\frac{\eta\lambda}{2}$. Then we come to our \textbf{first main} theorem:
\begin{theorem}
Given cost function $c(w)=\frac{\lambda}{2}\|{w}\|^2+L(x,y,\hat y,\widetilde x)$, $u(\cdot)$ the derivative function of $L$ over $\hat{y}$ with Lipschitz constant $c^*$. If learning rate $\eta$ and matrix element-wise bound is small enough, the overall mapping of Feras model converges  with contraction constant $1-\frac{\eta\lambda}{2}$.
\label{theorem1}
\end{theorem}

\subsection{Convergence rate and accuracy}
\label{convergence rate and accuracy}
\begin{theorem}
For any $z\in\mathbb N$, if $p^*$ is a contraction mapping on $(M,d)$ with contraction rate $(1-\varepsilon)$ and $D^*$ its fixed point. If the initial parameter distribution $D^0$ satisfies $W_z(D^0,D^*)\leq K$, then $W_z(D^T,D^*)\leq K(1-\varepsilon)^T$.
\label{theorem2}
\end{theorem}
The theorem above is given by paper \cite{NIPS2010_abea47ba}. By applying Theorem \ref{theorem1}, Theorem \ref{theorem2} and investigating its constraints in Appendix \ref{appendix B}, we find out the \textbf{main advantage} of our \textbf{Feras}  model:
\begin{restatable}[]{theorem}{restateb}
\label{theorem3}
If $\mathcal G^{n,t}(\mathcal V^{n,t},\mathcal E^{n,t})$ is a strongly connected graph for each host $n$ at each iteration $t$, for a given convergence rate, Feras model offers a looser constraint than simple shared weight GCN on element-wise bound of weight matrices.
\end{restatable}

Similar to a Parallelized Stochastic Gradient Descent Algorithm in paper \cite{NIPS2010_abea47ba}, the distance between final parameter distribution $D^T$ and minimizer of the cost function ($\ref{cost function}$) can be bounded.

\textbf{Remark: }Suppose that weight matrices are shared every $q$ iterations, its convergence rate is still maintained. In fact, as what Lemma \ref{lemma2} implies, note $p'_n=\phi^n \circ \cdots \circ \phi^n(w^n)$ ($q$ times of composition), $ (p^{*})'=\sum_{n=1}^{N}p'_n$ is still a contraction. Implementing into Theorem \ref{theorem2}, $(1-\varepsilon')=(1-\varepsilon)^p=(1-\frac{\eta\lambda}{2})^q$, then $W_z(D^T, D^*)\leq K(1-\varepsilon)^{T-\lfloor \frac{T}{q}\rfloor q}(1-\varepsilon')^{\lfloor \frac{T}{q}\rfloor}=K(1-\varepsilon)^T$ still holds.

\section{Experiments}
\subsection{Dataset discription}
\label{dataset}
We use three datasets to train our \textbf{Feras}  model: PPI, Flickr, and Reddit. Information on these datasets is summarized in table \ref{datasettable}. The protein-protein interaction (PPI) dataset is a collection of human tissue protein-protein association networks where edges represent direct (physical) protein-protein interactions or indirect (functional) associations between proteins. Flickr and Reddit are social networks correspondingly built by users of Flickr and Reddit. Flickr forms edges between images with the same properties such as geographic location, submitted gallery, common tags, etc. Reddit is a post-post graph network where two posts are connected if they are commented by the same user. Multi-classification tasks are performed on PPI while single classification tasks are performed on Flickr and Reddit.

As mentioned before, we adopt GraphSAINT as the sampler whose code is available in open source \footnote{\url{https://github.com/GraphSAINT/GraphSAINT}}. We adopt the same dataset as GraphSAINT \cite{zeng2020graphsaint} to better compare model performance.  Note here that several mechanisms are applied in GraphSAINT, which are “-Node” for random node sampler; “-Edge” for random edge sampler; “-RW” for random walk sampler; “-MRW” for multi-dimensional random walk sampler.

For comparison, we propose two types of baselines: 1. GraphSAINT with different sampler mechanisms ("-Node", "-Edge", "-RW", "-MRW"), hosts perform task in an isolated manner, i.e. neither embedding nor weight matrices are shared; 2.GraphSAINT-SW, simple federated learning is implemented to share weight matrices with the best sampler mechanism among "-Node", "-Edge", "-RW", and "-MRW" for different host numbers.
\begin{table}[htbp]
\caption{Dataset description}
  \centering
\begin{tabular}{ccccccc} 
        \toprule
        Dataset & Nodes & Edges & Degree & Feature & Classes & Train/Val/Test\\
        \midrule
        PPI & 14,755 &225,270 &15 &50 &121 &0.66/0.12/0.22\\
        Flickr & 89,250 &899,756 & 10 &500 &7 &0.50/0.25/0.25\\
        Reddit &232,965 &11,606,919 &50&602 &41 &0.66/0.10/0.24\\
        \bottomrule
\end{tabular}
\label{datasettable}
\end{table}

\subsection{Experimental setup}
In our application scenario, host number $N$ and proportion of private nodes $\pi$ are crucial variables. In our experiments, privates nodes are uniformly distributed to all hosts, and public nodes are approachable to everyone. $\kappa = \frac{N-1}{N}\pi$ of nodes are thus invisible to each host. What's more, we recall that $q$ defines sharing weight frequency, i.e. weight matrices are shared every $q$ epochs.

All codes are run in a tensorflow\footnote{\url{https://www.tensorflow.org/}} environment on NVIDIA Titan X (Pascal) GPU\footnote{\url{https://www.nvidia.com/en-us/geforce/products/10series/titan-x-pascal/}}. The whole network contains $3$ layers with first two convolutional layers and a final dense layer. Each layer employs a \textit{ReLU} as activation function. Embeddings are shared after the first layer. As we remarked in Section \ref{convergence rate and accuracy}, the size of $q$ brings no influence on convergence rate, we set $q=10$, and this is further clarified later.

Uncertain data transmission factors in federated learning will affect the experimental results, hence we employ only a single server to simulate the condition. More precisely, hosts are set to train the model sequentially instead of in parallel and an embedding sharing table of node corpus is created to simulate the AS. Within one epoch, former hosts finishing one iteration update the embedding sharing table, and latter hosts use the latest embedding table to train the model. Therefore, we guarantee that each host has access to the latest node embedding before their training iteration and analysis on simulation results can lead to our \textbf{Feras}  model performance in a stable data transmission environment.
\begin{table}[htbp]
\caption{Comparison of test set score with baseline methods}
  \centering
\begin{tabular}{ ccccc } 
\toprule
\multicolumn{2}{c}{} &PPI &Flickr &Reddit\\
\midrule
\midrule
\multirow{8}{4em}{$\kappa=40\%$} &GraphSAINT-Node&$0.867\pm 0.002$ &$0.496\pm 0.003$ &$0.957\pm 0.001$ \\
&GraphSAINT-Edge &$0.941\pm 0.002$ &$0.499\pm 0.004$ &$0.962\pm 0.002$\\
&GraphSAINT-RW &\bm{$0.952\pm 0.003$} &$0.500\pm 0.004$ &$0.962\pm 0.001$\\
&GraphSAINT-MRW &$0.922\pm 0.002$ &\bm{$0.501\pm 0.004$} &\bm{$0.959\pm 0.002$}\\
&GraphSAINT-SW ($N=3$) &$0.933\pm 0.002$ &$0.499\pm 0.001$ &$0.952\pm 0.002$\\
&GraphSAINT-SW ($N=5$) &$0.936\pm 0.003$ &$0.498\pm 0.002$ &None\footnotemark[4] \\
\cline{2-5}
&Feras ($N=3$) &\bm{$0.953\pm 0.003$} &\bm{$0.501\pm 0.002$} &\bm{$0.962\pm 0.002$}\\
&Feras ($N=5$) &$0.944\pm 0.002$ &$0.497\pm 0.002$ &None\footnotemark[4] \\
\midrule
\multirow{8}{4em}{$\kappa=60\%$} 
&GraphSAINT-Node &$0.740\pm 0.006$ &$0.492\pm 0.006$ &$0.950\pm 0.003$\\
&GraphSAINT-Edge &$0.814\pm 0.003$ &$0.496\pm 0.004$ &\bm{$0.959\pm 0.001$}\\
&GraphSAINT-RW &\bm{$0.819\pm 0.005$} &$0.493\pm 0.005$ &$0.958\pm 0.002$\\
&GraphSAINT-MRW &$0.809\pm 0.004$ &$0.495\pm 0.005$ &$0.956\pm 0.002$\\
&GraphSAINT-SW ($N=3$) &$0.783\pm 0.010$ &\bm{$0.497\pm 0.002$} &$0.955\pm 0.001$\\
&GraphSAINT-SW ($N=5$) &$0.809\pm 0.008$ &$0.494\pm 0.005$ &None\footnotemark[4]\\
\cline{2-5}
&Feras ($N=3$) &\bm{$0.895\pm 0.008$} &\bm{$0.499\pm 0.002$} &\bm{$0.960\pm 0.001$}\\
&Feras ($N=5$) &$0.872\pm 0.004$ &$0.496\pm 0.001$ &None\footnotemark[4] \\
\bottomrule
\end{tabular}
\label{comparison}
\end{table}

\begin{figure*}
\centering
\begin{subfigure}[h]{\linewidth}
\includegraphics[width=\linewidth]{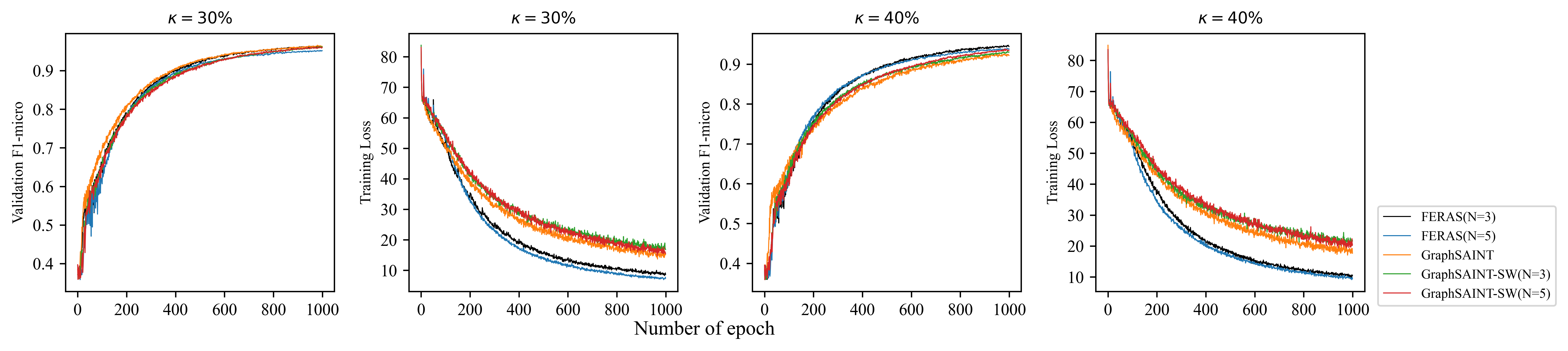}
\caption{Comparison for low $\kappa$ values ($30\%,40\%$)}
\end{subfigure}

\begin{subfigure}[h]{\linewidth}
\includegraphics[width=\linewidth]{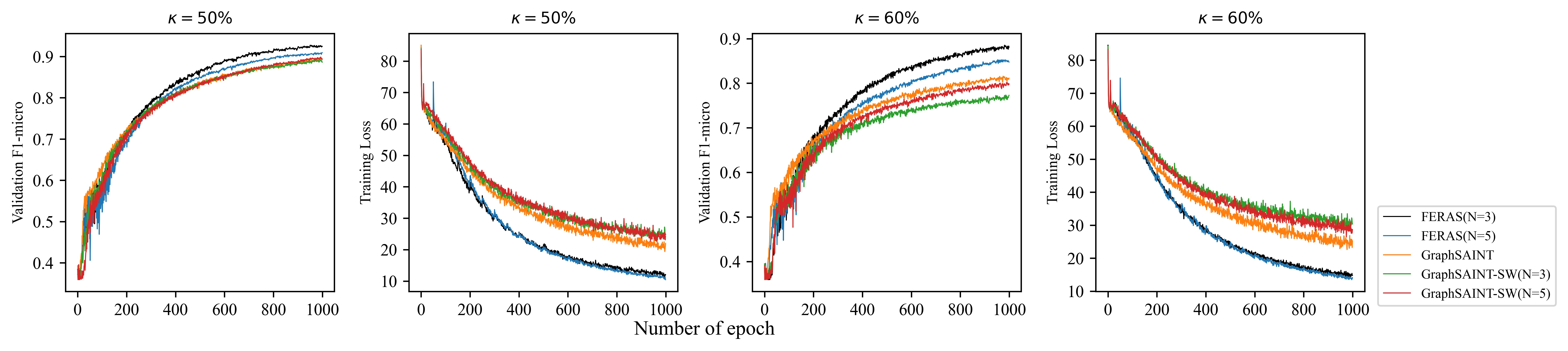}
\caption{Comparison for high $\kappa$ values ($50\%,60\%$)}
\end{subfigure}
\caption{Convergence curves of Feras model and baselines on PPI dataset}
\label{curves}
\end{figure*}

\subsection{Comparison with baselines}
\footnotetext[4]{The codes throw runtime error due to too large data size.}
We use the micro-averaged F1-micro score to measure program performance. Table \ref{comparison} provides an overview of F1 score comparison between \textbf{Feras}  and baseline models under different $\kappa$ and $N$. As well as baseline GraphSAINT-SW, we pick the finest sampler for \textbf{Feras} after trying all four different mechanisms ("-Node", "-Edge", "-RW", "-MRW"). The hidden dimension is kept to be the same across all methods. The mean and confidence interval of the accuracy values are measured by more than three rounds under the same hyperparameters. What's more, for experiments having more than one host, the final score is the mean of all hosts' score. 

Clearly, table \ref{comparison} shows that \textbf{Feras} achieves a significant improvement of accuracy on PPI and Reddit dataset and a slight advantage on Flickr dataset. Comparing two types of baselines defined in Section \ref{dataset}, one can find in table \ref{comparison} that the positive influence brought by simple implementation of federated learning is very limited (baseline 2), and sometimes it works even worse than isolated training (baseline 1). We can thus confirm that original Graph-based networks are not directly applicable to federated learning, or with a performance trade-off. This result may be explained by the fact that the existence of private nodes perturbs the derivatives of weight matrices and this bias is overlaid through weight matrix sharing. However, \textbf{Feras} is capable to step over this problem because the influence of invisible nodes is minimized via shared embeddings. 

\subsubsection{Discussion on influence of $\kappa$}
As $\kappa$ denotes the proportion of unseen nodes of each host, we further give convergence curves in Figure \ref{curves} to illustrate the performance of \textbf{Feras} under different $\kappa$ values on PPI dataset. The legend "GraphSAINT" is used to represent the best method of baseline 1, and the other notations stay the same as mentioned before. As can be seen from the figure, regarding $\kappa = 30\%$, \textbf{Feras} does not have a huge superiority of F1-micro score, when $\kappa$ raises to $40\%$, the advantage of \textbf{Feras} appears and becomes evident with $\kappa=50\%$, this advantage is further enhanced by $\kappa =60\%$. In terms of host number, since our algorithm is actually an optimization of the loss function, it is comprehensible that \textbf{Feras}($N=5$) converges slightly faster than \textbf{Feras}($N=3$), which can be implied in Appendix \ref{appendix B}. One interesting finding is that \textbf{Feras} always has a faster convergence rate of F1-score than baseline models under the same conditions, and this is an extension of looser constraint on element-wise bound of weight matrices for a given convergence rate mentioned in Theorem \ref{theorem3}. 

\begin{figure}[h]
  \centering
  \includegraphics[width=0.6\linewidth]{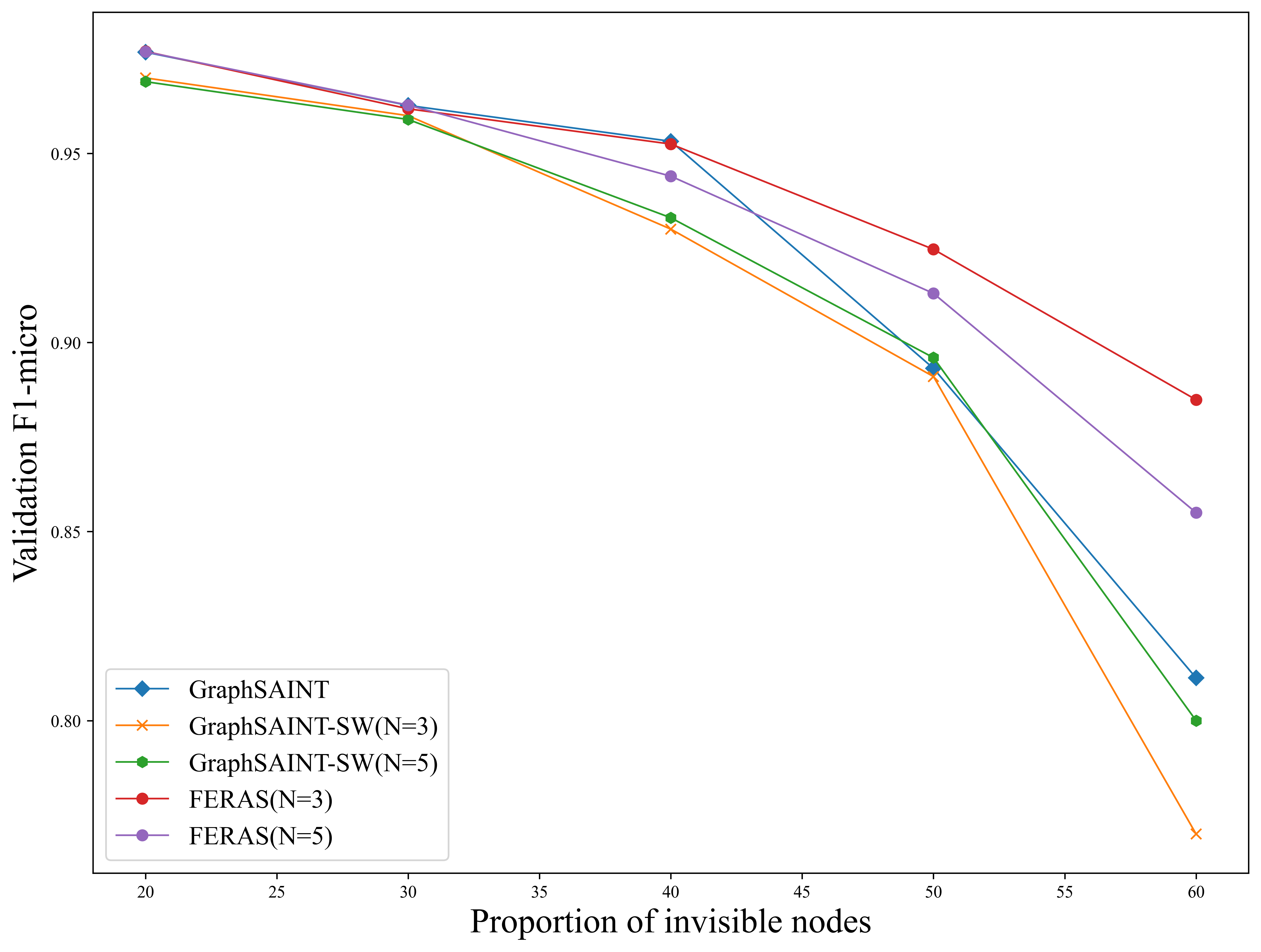}
  \caption{Sensitivity analysis of $\kappa$ on PPI dataset}
    \label{kval}
\end{figure}

Figure \ref{kval} provides a direct trend of accuracy along with increment of $\kappa$ on PPI dataset. 
It is not surprising to find out that less node attributes each host can see, more the accuracy decreases for all models. However, it is important to report that \textbf{Feras} can much better retain its performance than other baseline models and this advantage is magnified for higher $\kappa$. Again this result reveals the gravity to share embedding in a multi-host privacy scenario.

\subsubsection{Discussion on influence of $q$} 
\begin{figure}[h]
  \centering
  \includegraphics[width=0.6\linewidth]{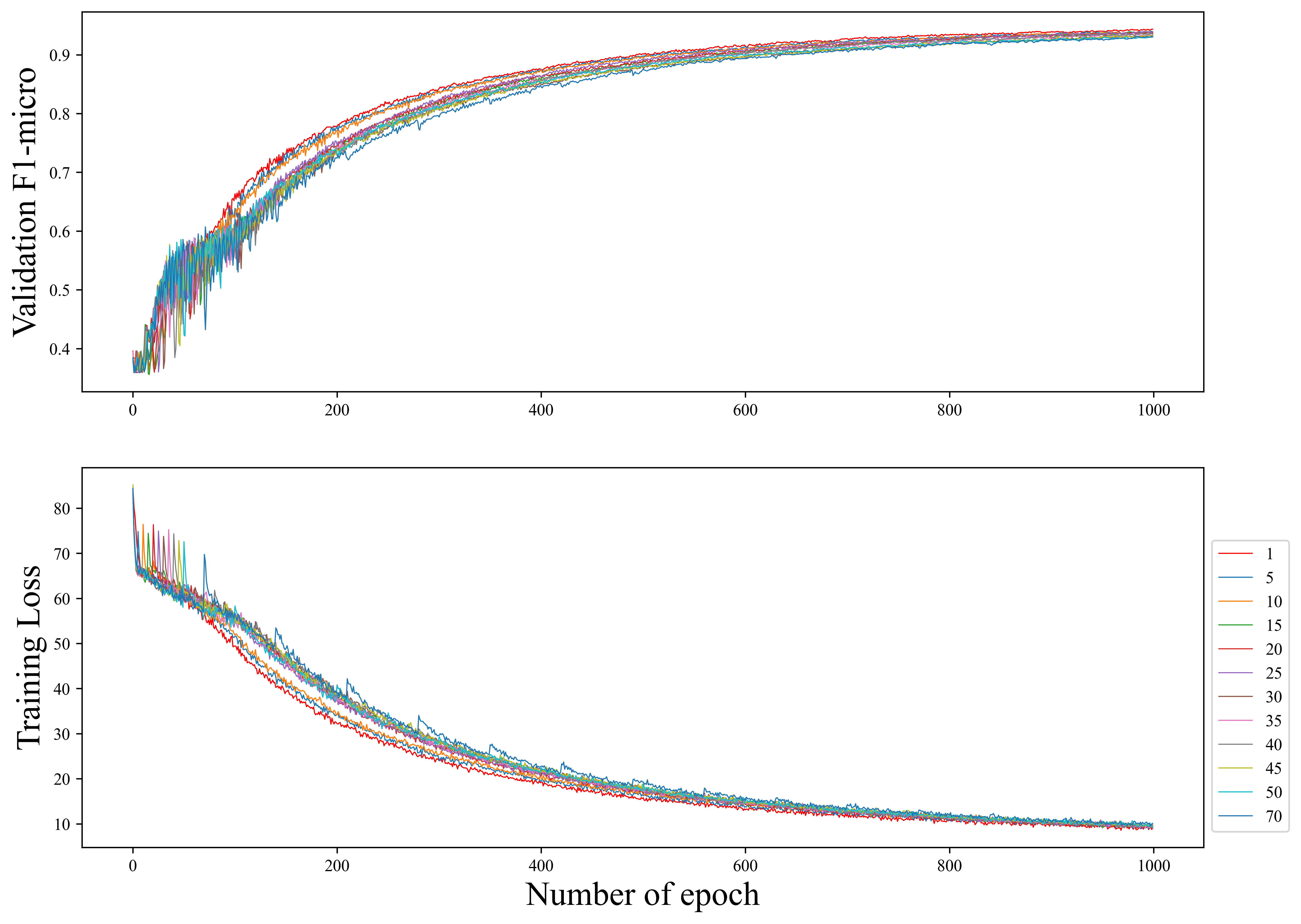}
  \caption{Sensitivity analysis of $q$ on PPI dataset}
    \label{qval}
\end{figure}

Figure \ref{qval} provides the convergence curves for different $q$ values. 
Experiments are conducted on PPI dataset, $N=5$, $\kappa=40\%$ and $q$ varies from 1 to 70. Coherent with what we proved in Section \ref{convergence rate and accuracy}, the convergence rate for models with different $q$ is almost the same. Since the performance of $q=10$ is very close to that of $q=1$ and $q=5$ between epoch 50 and epoch 400, while there is a significant gap between $q=10$ and $q=15$ over the same interval, considering information transmission costs and network delay, our choice of taking $q=10$ for previous experiments is justified.

\section{Conclusion}
In this paper, we have proposed a novel \textbf{Feras} model to solve graph-based networks training problem with private data. The core idea is to facilitate the sharing of node embeddings and weight matrices between hosts. Both theoretical and empirical methods have been adopted to evaluate our model. In theoretical aspect, we have proved the convergence of \textbf{Feras} and further compared its constraints with simple shared weight GCN, which offers a solid support for our model. In terms of experiments, a simulation system has been designed to run \textbf{Feras} on different datasets, we also interrogated the influence of the proportion of invisible nodes ($\kappa$) and the sharing weight frequency ($q$), all the experiments above have confirmed the advantage of our \textbf{Feras} model on convergence rate and accuracy. 

At present, we have only studied the training on uniformly distributed private nodes. In the future, heterogeneous data distribution can be investigated. What's more, our method focuses on privacy scenario, thus it is worth bringing forward attacks generated to leak user information and propose possible defensive mechanisms in the future.

\bibliographystyle{unsrt}  
\bibliography{references}  
\appendix

\section{Proof for Lemma \ref{lemma1}}
\label{appendix A}
Equation (\ref{cost function}), (\ref{update rule})  shows the necessity to derive the form of $\frac{\partial c(w^{n,,t}_i)}{\partial w^{n,t}_i}$ from $\frac{\partial \hat y^{n,t}}{\partial w^{n,t}}$, the Lipschitz constant of Equation (\ref{update rule}) thus depends exclusively on the Lipshitz constant of the feed forward network. Thanks to the work presented in \cite{combettes2020lipschitz}, the influence of non-expansive averaged activation functions can be neglected. Hence in the following discussion, we set our activation operator as a \textit{ReLU} function.

Unlike traditional neural networks, our model involves a shared-embedding layer. Thus we start by formulating $\hat y^{n,t}$ as a standard linear function of $w^{n,t}_i$, followed by giving explicit expression of $\frac{\partial \hat y^{n,t}}{\partial w^{n,t}}$, and then end with Lipschitz constant analysis. In this part, we only consider the parameter propagation for a given iteration $t(1\leq t\leq T)$, we don't specify $t$ to simplify the notation.

\subsection{Standard linear function}
\begin{corollary}
For each host $n$, $vec(\hat y^{n})$ can be expressed as a linear function of $w_i^n (i\in\{1,2\})$ in the form of
\begin{equation}
    vec(\hat{y})=\mathcal B^n_i+\mathcal M^n_ivec(w_i^n).
    \label{linear function}
\end{equation}
\label{corol 1}
\end{corollary}
\begin{proof}
Equation \ref{linear function} holds naturally for $i=2$ since $\hat y^n=\overline A^n\widetilde x^nw_2^n$:
\begin{equation}
    \mathcal B^n_2=0,\mathcal M_2^n=I_{k_n}\bigotimes (\overline A^n\widetilde x^n)
    \label{equa 5}
\end{equation}
Turning now to $i=1$, rearranging Equation (\ref{hat x}), (\ref{widetilde x}), (\ref{hat y}) and (\ref{connect equa}) yields:
\begin{align}
\hat y^n=\overline A^n\widetilde x^nw_2^n =\overline A^n\Theta^n\hat xw_2^n=\overline A^n\Theta^n(\hat x_{\mathcal V-\mathcal V^n}+\hat x_{\mathcal V^n})w_2^n
\label{equa 1}
\end{align}
where $\hat x_{\mathcal V^n}$ is obtained by fetching elements belonging to $\mathcal V^n$ from $\hat x$, i.e.:
\begin{equation}
    \hat x_{\mathcal V^n}=\begin{pmatrix}0\\\vdots\\0\\\hat x^n\\0\\\vdots\\0\end{pmatrix}=\begin{pmatrix}0\\\vdots\\0\\\overline A^nx^nw_1^n\\0\\\vdots\\0\end{pmatrix}=\begin{pmatrix}0\\\vdots\\0\\\overline A^n\\0\\\vdots\\0\end{pmatrix}x^nw_1^n
\end{equation}
Similarly, we define:
\begin{equation}
    \hat x_{\mathcal V-\mathcal V^n}=(\hat x^1\cdots \hat x^{n-1}\quad 0\cdots 0\quad \hat x^{n+1}\cdots \hat x^N)^T
\end{equation}
Therefore, Equation (\ref{equa 1}) can be reformulated as :
\begin{equation}
    \hat y^n=\overline A^n\Theta^n\hat x_{\mathcal V-\mathcal V^n}w_2^n +\overline A^n\Theta^n (0 \cdots 0 \overline A^n 0 \cdots 0)^Tx^nw_1^nw_2^n
    \label{equa 2}
\end{equation}
Since for all matrix product $Y=AXB$, $vec(Y)=(B^T\bigotimes A)vec(X)$.  Defining $\mathcal B^n_1$, $\mathcal M^n_1$ as following :
\begin{align}
    &\mathcal B^n_1=vec(\overline A^n\Theta^n\hat x_{\mathcal V-\mathcal V^n}w_2 ^n),\\
    &\mathcal M^n_1=((N^n)^T\bigotimes M^n) \label{equa 11}
\end{align}
where $ M^n=\overline A^n\Theta^n(0 \cdots 0 \overline A^n 0 \cdots 0)^Tx^n, N^n=w_2^n$, Equation (\ref{equa 2}) can be further expressed in a standard linear way :
\begin{equation}
    vec(\hat y^n)=\mathcal B^n_1+\mathcal M^n_1vec(w_1^n), \mathcal M_1^n\in \mathbb R^{m_3k_n\times m_1m_2}
\end{equation}
\end{proof}

\begin{corollary}
Given cost function $c(w)=\frac{\lambda}{2}\|{w}\|^2+L(x,y,\hat y,\widetilde x)$ and $u(\cdot)$ the derivative function of $L$ over $\hat{y}$, for each host $n$ and $i\in\{1,2\}$, the update rule  $\phi(w^n):=w^{n}-\eta\nabla c(w^n)$ allows a formal solution :
\begin{equation}
    vec(\phi(w_i^n))=(1-\eta\lambda)vec(w_i^n)-\eta(\mathcal M^n_i)^Tu(\mathcal B^n_i+\mathcal M^n_i vec(w_i^n))
    \label{equa 4}
\end{equation}
\label{corol 2}
\end{corollary}
\begin{proof}
Chain rule shows that 
\begin{equation}
    vec(\phi(w_i^n))=(1-\eta\lambda)vec(w_i^n)-\eta\frac{\partial vec(\hat y^n)}{\partial vec(w_i^n)}\frac{\partial }{\partial vec(\hat y^n)}L(\hat y^n,y^n)|_{\hat y^n}    
\end{equation}
What's more, Corollary \ref{corol 1} shows that $\Delta vec(\hat y^n)=\mathcal M^n_i\Delta vec(w_i^n)$. Consequently, we obtain:
\begin{align}
    vec(\phi(w_i^n))&=(1-\eta\lambda)vec(w_i^n)-\eta(\mathcal M^n_i)^Tu(vec(\hat y^n))\\&=(1-\eta\lambda)vec(w_i^n)-\eta(\mathcal M_i^n)^Tu(B^n_i+\mathcal M_i^nvec(w_i^n))
\end{align}
\end{proof}

\subsection{Lipschitz constant analysis}
\label{subappendix A}
\begin{lemma}
Given cost function $c(w)=\frac{\lambda}{2}\|{w}\|^2+L(x,y,\hat y,\widetilde x)$ and $u(\cdot)$ the derivative function of $L$ over $\hat{y}$, $u(\cdot)$ is Lipschitz continous with constant $c^*$. If the spectral radius $\rho((\mathcal M_1^n)^T\mathcal M_1^n)\leq \frac{\lambda}{2c^*}$ and $\eta\leq\frac{4}{2c^*\rho((\mathcal M_2^n)^T\mathcal M_2)+3\lambda}$, for each host $n$ and $i\in\{1,2\}$, the update rule  $\phi(w^n):=w^{n}-\eta\nabla c(w^n)$ is also Lipschitz continuous with constant $1-\frac{\eta\lambda}{2}$ and is thus a contraction.
\label{corol 3}
\end{lemma}
\begin{proof}

The conclusion drawn in Corollary \ref{corol 2} reveals that update rule can be represented by the sum of two Lipshitz function, whose Lipschitz constants are respectively $1-\eta\lambda$ and $\|c^*\eta(\mathcal M_i^n)^T\mathcal M_i^n\|_2$. We develop our proof by looking into weight matrices of different layers.

With regard to $i=1$, for any two different weight matrices $v_1^n$ and $w_1^n$, the following inequality holds: 
\begin{align}
&\|vec(\phi(w_1^n))-vec(\phi(v_1^n))\|\leq (1-\eta\lambda+\|c^*\eta(\mathcal M_1^n)^T\mathcal M_1^n\|_2)\|w_1^n-v_1^n\|
\label{equa 3}
\end{align}
Knowing that $\|(\mathcal M_1^n)^T\mathcal M_1^n\|_2\leq \rho((\mathcal M_1^n)^T\mathcal M_1^n)\leq \frac{\lambda}{2c^*}$, Equation (\ref{equa 3}) can be stated as :
\begin{equation}
    \|vec(\phi(w_1^n))-vec(\phi(v_1^n))\|\leq (1-\frac{\eta\lambda}{2})\|w_1^n-v_1^n\|
    \label{equa 12}
\end{equation}

Moving now to $i=2$, Equation (\ref{equa 4}) shows that :
\begin{align}
&vec(\phi(w_2^n))-vec(\phi(v_2^n))=(1-\eta\lambda)[vec(w_2^n)-vec(v_2^n)]-\notag\\&\eta(\mathcal M_2^n)^T[u(\mathcal B_2^n+\mathcal M_2^n vec(w_2^n))-u(\mathcal B_2^n+\mathcal M_2^n vec(v_2^n))]
\label{equa 6}
\end{align}
Assuming $w_2^n\geq v_2^n$, since $L$ is a convex function, $u(\cdot)$ is increasing, $\mathcal M_2^n$ from Equation (\ref{equa 5}) is nonnegative. So :
\begin{equation}
    vec(\phi(w_2^n))-vec(\phi(v_2^n))\leq(1-\eta\lambda)[vec(w_2^n)-vec(v_2^n)]
    \label{equa 7}
\end{equation}
Noticing $u(\cdot)$ a Lipschitz continous function :
\begin{align}
    &vec(\phi(w_2^n))-vec(\phi(v_2^n))\geq \notag \\&[(1-\eta\lambda)-\eta c^*(\mathcal M_2^n)^T\mathcal M_2^n](vec(w_2^n)-vec(v_2^n))
    \label{equa 8}
\end{align}
What's more, 
\begin{align}
\|[(1-\eta\lambda)-\eta c^*(\mathcal M_2^n)^T\mathcal M_2^n](vec(w_2^n)-vec(v_2^n))\|_F\notag\\\leq\|(1-\eta\lambda)-\eta c^*(\mathcal M_2^n)^T\mathcal M_2^n\|_2\|vec(w_2^n)-vec(v_2^n)\|_F
\label{equa 9}
\end{align}
$(\mathcal M_2^n)^T\mathcal M_2^n$ is a real symmetric matrix, there exists an unitary matrix $Q$ such that $(\mathcal M_2^n)^T\mathcal M_2^n=Q\Lambda Q^T$ where $\Lambda$ is characterised by its diagonal elements $\lambda(\Lambda)=(\lambda_1,\lambda_2,\cdots,\lambda_{k_nm_2})$:
\begin{align}
\|(1-\eta\lambda)-\eta c^*(\mathcal M_2^n)^T\mathcal M_2^n\|_2&=\|Q((1-\eta\lambda)I-\eta c^*\Lambda)Q^T\|_2\notag\\&=\rho((1-\eta\lambda)I-\eta c^*\Lambda)
\end{align}
The condition $\eta\leq\frac{4}{2c^*\rho((\mathcal M_2^n)^T\mathcal M_2)+3\lambda}$ implies that no matter $\eta c^*\rho(\Lambda)\leq(1-\eta\lambda)$ or not, 
\begin{align}
   \rho((1-\eta\lambda)I-\eta c^*\Lambda)\leq 1-\frac{\eta\lambda}{2} 
   \label{equa 10}
\end{align}
Rearranging Equation (\ref{equa 7}), (\ref{equa 8}), (\ref{equa 9}) and (\ref{equa 10}), we get :
\begin{equation}
    \|vec(\phi(w_2^n))-vec(\phi(v_2^n))\|\leq (1-\frac{\eta\lambda}{2})\|w_2^n-v_2^n\|
    \label{equa 13}
\end{equation}

Putting Equation (\ref{equa 12}), Equation (\ref{equa 13}) and Definition \ref{Contraction} together, we can conclude that the update rule is a contraction.
\end{proof}
Now taking a look at the composition of $\mathcal{M}_1^n$ in Equation (\ref{equa 11})and $\mathcal{M}_2^n$ in Equation (\ref{equa 5}), it's evident that the constraints mentioned in Lemma \ref{corol 3} concern actually only $\eta$ and $w_2^n$.

\section{Proof for Theorem \ref{theorem3}}
\label{appendix B}
Theorem \ref{theorem1} shows that the parameter convergence rate  depends mainly on the contraction constant of the overall model mapping, the proof of Theorem \ref{theorem2} in Appendix \ref{subappendix A} gives two constraints for a given contraction constant:
\begin{align}
    &\eta\leq\frac{4}{2c^*\rho((\mathcal M_2^n)^T\mathcal M_2)+3\lambda} \label{cnstr 1}\\
    &\rho((\mathcal M_1^n)^T\mathcal M_1^n)\leq \frac{\lambda}{2c^*} \label{cnstr 2}
\end{align}

Constraint (\ref{cnstr 1}) relies on $\mathcal{M}_2^n$, while Equation (\ref{equa 5}) shows the dependence of $\mathcal{M}_2^n$ on graph topological structure and input vector, which is fixed in each iteration. However, constraint (\ref{cnstr 2}) contains a more interesting intension.

Equation (\ref{equa 11}) shows that:
\begin{align}
    \lambda((\mathcal M_1^n)^T\mathcal M_1^n)=&\lambda((N\bigotimes M^T)(N^T\bigotimes M))\notag\\=&\lambda(NN^T\bigotimes M^TM)\notag\\=&\{\mu_i\delta_j,\mu_i\in\lambda(NN^T),\delta_j\in\lambda(MM^T)\}\notag
\end{align}
\begin{equation}
\Rightarrow \qquad \rho(\mathcal{M}_1^n)^T\mathcal M_1^n=\rho(NN^T)\rho(MM^T)=\rho(w_2^n(w_2^n)^T)\rho(MM^T)   
\label{equa 14}
\end{equation}

Use $\underline{MM^T}$ instead of $MM^T$ in case of general GCN without shared-embedding layer for distinction. Notice that
\begin{equation}
    \Theta^n(0 \cdots 0 \overline A^n 0 \cdots 0)^T=\Theta^{n,*}\overline A^n
\end{equation}
where $\Theta^{n,*}$ is a diagonal matrix and its diagonal elements are 
\begin{equation}
(\frac{1}{\widetilde{\mathcal{V}}^t.count(v^n_1)},\frac{1}{\widetilde{\mathcal{V}}^t.count(v^n_2)},\cdots,\frac{1}{\widetilde{\mathcal{V}}^t.count(v^n_{k_n})})    
\end{equation}
where
$\frac{1}{\widetilde{\mathcal{V}}^t.count(v)}=0$ if $\widetilde{\mathcal{V}}^t.count(v)=0 $

So :
\begin{align}
    &MM^T=\overline A^n\Theta^{n,*}\overline A^nx^n(\overline A^n\Theta^{n,*}\overline A^nx^n)^T\\
    &\underline{MM^T}=\overline A^nI_{k_n}\overline A^nx^n(\overline A^n I_{k,n}\overline A^nx^n)^T
\end{align}

$MM^T$ and $\underline{MM^T}$ are positive semi-definite, thus their eigenvalues are both nonnegative. What's more, they are both Hermitian matrices satisfying following property stated in \cite{enwiki:1036319874}:
\begin{corollary}
If $A$ is a Hermitian matrix of dimension $n$, for all $v\in\mathcal C^n$, $\|Av\|\leq \rho(A)\|v\|$ and $\| \cdot \|$ is the euclidean norm.
\end{corollary}
Since $\mathcal G^n$ is a stronly connected graph, $MM^T$ is an irreducible non-negative matrix. According to the Perron–Frobenius theorem \cite{doi:10.1137/S0036144599359449}, if $\lambda'$ is the dominant eigenvalue of $MM^T$,i.e.$|\lambda'|=\rho(MM^T)$, there exists always an eigenvector $X$ with eigenvalue $\lambda'$ whose components are all positive. Then:
\begin{align}
   \rho(MM^T)\|v\|=\|MM^Tv\|\leq\underline{MM^Tv}\leq\rho(\underline{MM^T})\|v\| 
\end{align}

Now we obtain $\rho(MM^T)\leq\rho(\underline{MM^T})$, applying it to Equation (\ref{equa 14}) and constraint \ref{cnstr 2}, a looser constraint with shared-embedding layer is proven.

\end{document}